\documentclass{article}

\usepackage[nonatbib,final]{neurips_2022}

\usepackage[utf8]{inputenc} % allow utf-8 input
\usepackage[T1]{fontenc}    % use 8-bit T1 fonts
\usepackage{hyperref}       % hyperlinks
\usepackage{url}            % simple URL typesetting
\usepackage{booktabs}       % professional-quality tables
\usepackage{amsfonts}       % blackboard math symbols
\usepackage{nicefrac}       % compact symbols for 1/2, etc.
\usepackage{microtype}      % microtypography
\usepackage{xcolor}         % colors

\usepackage{amsmath,amssymb,amsfonts}
\usepackage{graphicx}
\usepackage{textcomp}
\usepackage{xcolor}
\usepackage{subcaption}
\usepackage{amsmath,amssymb}
\usepackage{amsthm}
\usepackage{algorithm}
\usepackage{dsfont}
\usepackage{multirow}

\usepackage{wrapfig}
\usepackage{algorithm} 
\usepackage{algpseudocode} 

\usepackage[disable]{todonotes}

\newcommand{\chugo}[1]{\todo[color=red!60,inline]{H:#1}}

\renewcommand{\figurename}{Fig.}
\newcommand{\bgi}{{\sc bgi}}

\title{Pragmatically Learning from Pedagogical Demonstrations in Multi-Goal Environments}

\author{%
  Hugo Caselles-Dupré, Olivier Sigaud, Mohamed Chetouani \\
  Sorbonne Université, CNRS, Institut des Systèmes Intelligents et de Robotique (ISIR) \\
  Paris, France\\
  \texttt{casellesdupre.hugo@gmail.com,olivier.sigaud,mohamed.chetouani@isir.upmc.fr} \\
}

\begin{document}

\maketitle

\begin{abstract}
Learning from demonstration methods usually leverage close to optimal demonstrations to accelerate training. By contrast, when demonstrating a task, human teachers deviate from optimal demonstrations and pedagogically modify their behavior by giving demonstrations that best disambiguate the goal they want to demonstrate. Analogously, human learners excel at pragmatically inferring the intent of the teacher, facilitating communication between the two agents. These mechanisms are critical in the few demonstrations regime, where inferring the goal is more difficult. In this paper, we implement pedagogy and pragmatism mechanisms by leveraging a Bayesian model of Goal Inference from demonstrations (\bgi). We highlight the benefits of this model in multi-goal teacher-learner setups with two artificial agents that learn with goal-conditioned Reinforcement Learning. We show that combining \bgi-agents (a pedagogical teacher and a pragmatic learner) results in faster learning and reduced goal ambiguity over standard learning from demonstrations, especially in the few demonstrations regime. We provide the code for our experiments \footnote{\url{https://github.com/Caselles/NeurIPS22-demonstrations-pedagogy-pragmatism}}, as well as an illustrative video explaining our approach \footnote{\url{https://youtu.be/V4n16IjkNyw}}.
\end{abstract}

%\section{Introduction}

%Imagine a teacher-learner setup where 3 colored blocks are lying on a table, the green and red blocks are close to each other and the blue block is further away. Assume the teacher wants to demonstrate a goal state where the blue block is close to the green one. A naive teacher may move the blue block to the green one, but this would be ambiguous: is the goal to put the blue block close to the green one or to the red one? By contrast, a pedagogical teacher would put the green block close to the blue one. To discover that this is the right demonstration, the pedagogical teacher may ask herself what goal she would infer if she was the student. This inference can be computed using its own policy: what is the probability of the right goal given a demonstration? 

%Now, the learner has to interpret the demonstration to predict the goal. Given the same inference mechanism as the teacher, a literal learner would perform this inference using its policy while a pragmatic learner would train its policy to predict its own goals from its own trajectories, increasing its likelihood to correctly infer the teacher's goal.

\section{Introduction}

\chugo{pourquoi est il interessant de donner des demos + goals, alors que les goals sont choisis de manière uniforme}

\chugo{faire une passe sur les termes de sciences cognitives, et se poser la question de si en ML on comprend ce qu'on dit}

\chugo{si on veut gagner de la place > goal conditioned policy abrégé en GCP}

\begin{figure}[ht]
    \centering
    \includegraphics[scale=0.57]{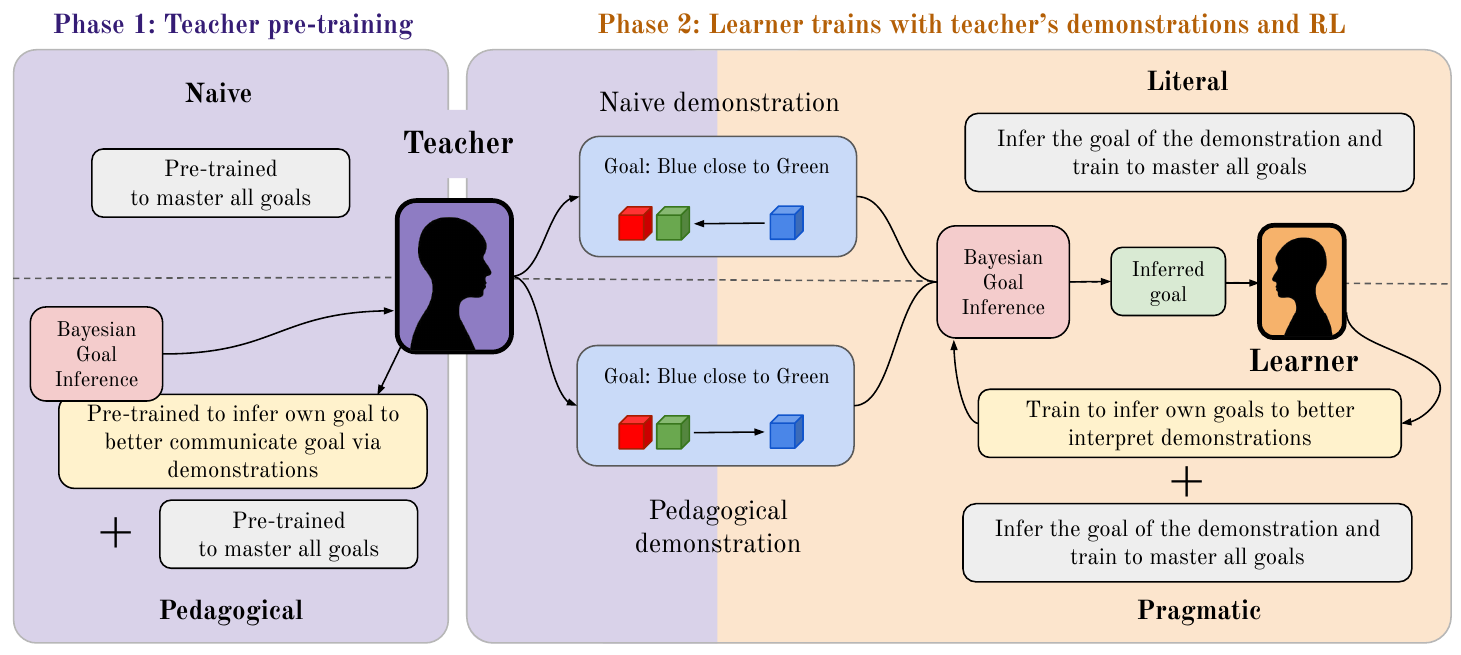}
    \caption{Overview of our teacher-learner setup. All goal inferences are performed using Bayesian Goal Inference, an inference method that computes the goal probability from a demonstration using the agent’s policy. In the first phase, the naive teacher is pre-trained to master all goals while the pedagogical teacher is additionally pre-trained to better infer goals from its own trajectories and effectively modifies its policy to produce demonstrations with less ambiguous goals. In the second phase, the learner is trained using the teacher’s demonstrations and goal-conditioned Reinforcement Learning. The literal learner infers goals from demonstrations and is rewarded for correct inferences while the pragmatic learner additionally trains itself to infer its own goals from its own trajectories and effectively modifying its policy to better infer goals from the teacher’s demonstrations.}
    \label{fig:main}
\end{figure}

Imagine a teacher-learner setup where 3 colored blocks are lying on a table, the green and red blocks are close to each other and the blue block is further away. Assume the teacher wants to demonstrate how to have the blue block close to the green one (see \figurename~\ref{fig:main}). A naive teacher may move the blue block to the green one, but this would be ambiguous: is the goal to put the blue block close to the green one or to the red one? By contrast, a pedagogical teacher would move the green block close to the blue one, hence away from the red one, resolving the ambiguity as illustrated in \figurename~\ref{fig:main}. 

As this example shows, in real life, when a teacher shows how to do something with a demonstration, the agent receiving the demonstration does not have access to the intended goal of the demonstration: it must infer this goal. In many situations, a single demonstration may be correctly interpreted as demonstration of a variety of goals – we call this goal ambiguity. In that case, the teacher may use pedagogy to help the learner infer the intended goal, and the learner may use pragmatism in order to increase its chance to infer the right goal. This is the situation we address in this paper.

To infer that a demonstration facilitates learning, the pedagogical teacher may ask herself what goal she would infer if she was the learner. 
%This inference can be computed using her own policy to infer the goal: as policies provide action probabilities given a state and a goal, it is possible to compute and sample goal probabilities from a distribution of demonstration probabilities over the goal space. Using her policy, the teacher can thus determine the probability that she would infer the right goal given this demonstration, and assume the learner would do the same.
Using her policy and Bayesian inference, she can determine the probability that she would infer the right goal given this demonstration, and assume the learner would do the same.
Now, the learner has to interpret the demonstration to infer the goal. Given the same inference mechanism as the teacher, a literal learner would perform this inference using her own policy. On top of that, a pragmatic learner could first train her policy to predict her own goals from her own trajectories, increasing her likelihood to correctly infer the teacher's goal.

Those mechanisms, pedagogy from the teacher and pragmatism from the learner, facilitate communication between the two agents and thus improve learning by reducing ambiguity about the inferred goal. Pedagogy and pragmatism are concepts borrowed from cognitive science research. On the one hand, pedagogy is defined as the optimization of teaching concepts from examples \cite{buchsbaum2011children, butler2014preschoolers}. On the other hand, pragmatism is a property used to resolve ambiguities of intention interpretation from teaching signals, which can be language with for instance the rational speech act (RSA) \cite{neale1992paul, goodman2016pragmatic}, or actions with pedagogical demonstrations \cite{shafto2014rational}.

Such ideas have been leveraged into the general theory of "Inferential Social Learning" (ISL) \cite{gweon2021inferential} in order to explain our abilities as learners and teachers to "interpret and generate evidence in social contexts". ISL is characterized as an inference mechanism guided by an intuitive understanding of how people think, plan, and act. It leverages Bayesian inference to explain several key aspects involved in pedagogy and pragmatism: pedagogy can help avoid goal ambiguity \cite{ho2019communication, ho2021cognitive} or accelerate goal inference from a learner (i.e. legibility) \cite{ho2016showing}, and pragmatism can help better infer goals \cite{falck2006infants}, or detect pedagogy \cite{gweon2021inferential}, or generalize from induction \cite{gweon2010infants} (what can I deduce from a teaching signal?). In this paper, we show that using Bayesian inference for disambiguating goals (pedagogy) and better infer goals (pragmatism) can improve the learning capabilities of artificial agents. In particular, we show that this mechanism is critical to reduce ambiguity when the teacher performs few demonstrations, which may contribute to more feasible robot teaching in the future.

%Using actions and languages as communication channels, parents and children respectively use pedagogy and pragmatism in the tutoring/learning process. Teachers might use instructions adapted to the learner, specific language that is easy to understand such as "motherese" \cite{gleitman1984current,saintgeorges2013,lim2014mei}, or exaggerated demonstrations and actions known as "motionese" \cite{nagai2007can, nagai2009computational, brand2002evidence}. Learners might use inductive generalization to infer what the teacher is trying to communicate, deducing relevant information from the way data is presented to them. These aspects are crucial for the efficiency of learning and teaching, and carefully studied in developmental psychology in order to better understand the cognitive upbringing of infants.

%Can artificial agents also benefit from implementing these social mechanisms? Artificial Intelligence (AI) and Robotics have recently made significant progress on behavior learning: on autonomous setups with retro video-games \cite{mnih2013playing}, manipulation \cite{rajeswaran2017learning} or control tasks \cite{salimans2017evolution}; or on interactive setups such as learning from demonstrations \cite{hester2018deep} or from descriptions and feedback \cite{colas2020language}. However, contrary to how humans teach and learn, most AI approaches do not take advantage of the social communication and mutual inferences that happen in human social learning. 

We present an approach for implementing inferential mechanisms to improve learning from demonstrations in an artificial teacher-learner setup, as illustrated in \figurename~\ref{fig:main}. We use two multi-goal environments: a simple one for illustrating our point, and a more complex simulated robotics block manipulation environment to show the robustness of our approach. In both environments, we create naive teachers trained to master all goals, and pedagogical teachers trained to provide demonstrations that best disambiguate between all goals and thus facilitate goal inference for learners. We then create literal learners, which learn from teacher demonstrations, and pragmatic learners, which adapt their policy to better infer goals from the teacher's demonstrations. The mechanism for implementing pedagogy and pragmatism is an additional reward for correct goal inference of the agent's own trajectories. Goal inference is implemented using a Bayesian model and the agent's policy.

We make the following contributions:

\begin{enumerate}
    \item Inspired by the ISL framework, we introduce pedagogical teaching and pragmatic learning mechanisms based on Bayesian Goal Inference and show that these mechanisms can be leveraged in a cooperative artificial teacher-learner training setup where both agents use goal-conditioned policies.
    \item We show that pedagogy and pragmatism help reduce goal ambiguity and result in faster learning, which is especially helpful when the teacher performs fewer demonstrations.
\end{enumerate}

\section{Related work}

Our work is related to several strongly connected areas.

\textbf{Bayesian inference.} Bayesian Inference was already used as a key mechanism for goal inference in the context of inverse planning \cite{baker2007goal, baker2008theory, zhi2020online}. It uses the Bayes formula to compute the probabilities of goals given the actions and the policy. In our work we use it as a tool to produce non-ambiguous demonstrations on the teacher side and pragmatic inference on the learner side.

\textbf{Pedagogical demonstrations.} In an attempt to explain how humans demonstrate tasks to each other, Ho et al. \cite{ho2016showing, ho2018effectively, ho2019communication} introduce the Observer Belief MDP model and show that producing more pedagogical demonstrations results in better performance. However, their work does not investigate how this affects an artificial learner, nor the effect of pragmatism in the learner.

\textbf{Legibility in Human-Robot Interaction.} Similarly to us, \cite{dragan2013legibility, lichtenthaler2011towards} modify robot action sequences to allow an observer (a human in their case) to more quickly and successfully understand a robot's goal from a more legible trajectory. 
%The intent expressive motion, termed legible, is obtained by maximizing inference as to what the goal of the action sequence might be by an observer model (transparency through motion). 
In our work, Bayesian goal inference is employed in pedagogical teachers to allow them to generate less ambiguous goal demonstration using additional reward for correct goal inference on their own motion.

\textbf{Pragmatic inference.} A long line of work in linguistics, natural language processing, and cognitive science has studied pragmatics: how linguistic meaning is affected by context and
communicative goals \cite{fried2018speaker, frank2012predicting, goodman2013knowledge}. Another line of work on pragmatics, more related to our work, focuses on inferring meaning from action in a context. It has been applied to Robotics \cite{milli2020literal, fisac2020pragmatic}, where a pragmatic robot better infers the objective of a teacher by considering its pedagogical intentions, in a multi-agent game theory context. In our case, we are interested in how a pragmatic learner can learn by itself to better infer the goals of a demonstrator, resulting in faster task learning, which is different from detecting if the teacher is pedagogical or not.

%\textcolor{blue}{\textbf{Multi-agent systems.} Reasoning about the other agents' actions, intentions and goals is central to Multi-Agent systems.}

%\textbf{Multi-agent systems.} Communication between agents has been shown to significantly improve efficiency, in particular with the centralized training and decentralized execution (CTDE) framework \cite{foerster2016learning,lowe2017multi,Filippos2020}. The nature of interaction between agents can either be cooperative, competitive or both. Inference of other agent’s goals is central to multi-agent reinforcement learning leading to the agents modeling agents approach \cite{HernandezLeal2019ASA}. In \cite{Raileanu18}, each agent explicitly models the behavior of the other agent using its own policy, while assuming that agents share a fixed set of goals and have similar abilities. In our approach, we do not model the behavior of the other. Teacher and learner exploit Bayesian goal inference to facilitate communication with each other in a multi-goal environment.

\textbf{Demonstrations in goal-conditioned tasks.} Our learning from demonstration mechanism is related to work on learning from demonstrations in multi-goal environments, by contrast with the more common setup where the agent only learns one task. Most approaches targeting fewer demonstrations  address the one task setup \cite{abbeel2004apprenticeship, ziebart2008maximum, zhang2018deep, stadie2019one}. Approaches using demonstrations for goal-conditioned tasks such as GAIL \cite{ho2016generative}, DCRL \cite{dance2021conditioned} and CLIC \cite{fournier2019clic} combine goal-conditioned RL and imitation learning with additional rewards. In our work, we also combine a goal-conditioned RL algorithm (GANGSTR \cite{akakzia2022help}) and a slightly modified version of an algorithm learning both from demonstration and from the agent's own experience (SQIL \cite{haarnoja2018soft}). The specificity of our pipeline is that we can easily improve it with pedagogical demonstrations and pragmatic inference to accelerate training, but in principle all goal-conditioned RL algorithms using demonstrations could be improved in such a way.

\section{Methods}

We consider multi-goal environments with teacher-learner scenarios where the learner is both trained with teachers' demonstrations and using its own exploration. The teacher $T$ and the learner $L$ are both represented by their respective goal-conditioned policies $\pi_T(.|g)$ and $\pi_L(.|g)$. These policies might be learned using any algorithm (multi-armed bandits, RL, evolution strategies, etc.), here we use a goal-conditioned RL (GCRL) algorithm \cite{colas2020intrinsically} called GANGSTR \cite{akakzia2022help}, which we describe in our experimental setup. Our multi-goal environments present goal ambiguity, as the same trajectory can simultaneously reach at least two goals $(g_1, g_2)$. Considering this hypothesis, it is generally not possible to reliably predict the pursued goal from a demonstration reaching both $g_1$ and $g_2$.

Both agents share common goal, state and action spaces. They communicate through teacher's demonstrations, learner's inferred goal and teacher feedback on this inference. To efficiently exploit these signals, we introduce a Bayesian Goal Inference ({\sc bgi}) mechanism helping agents infer goals from demonstrations. The teacher uses it to generate less ambiguous demonstrations (pedagogy), while the learner uses it to infer the teacher's goal from the demonstration (pragmatism). We refer to the pedagogical teacher and pragmatic learners as \bgi-agents.

The training process is spread across two phases: 1) the teacher is pre-trained to master all goals in the environment using GCRL, and 2) the learner infers the goal from a teacher's demonstration using Bayesian Goal Inference, the teacher provides feedback on the inference, and the learner is rewarded for correct predictions. The learner then attempts at reaching the inferred goal and iteratively improves its policy using GCRL.

Below we formally define {\sc bgi}, then we define naive teachers and literal learners, which do not leverage the full benefits of {\sc bgi}. Finally we introduce pedagogical teachers and pragmatic learners, which use {\sc bgi} to facilitate communication with each other.

\subsection{Bayesian Goal Inference in Teacher-Learner Interactions}
\label{eq:bgi}

We formally introduce the {\sc bgi} mechanism used to infer goals from a goal-conditioned policy. In our work, 1) the teacher uses it to implement pedagogy, 2) the learner uses it to infer goals from the demonstrations of the teacher, and 3) the learner uses it to implement pragmatism.

\paragraph{Inferring the Goal from a Demonstration}
\label{sec:bgi}

In a goal-conditioned Markov environment, the probability of observing a demonstration $d=((s_1, a_1), .., (s_n, a_n))$ given a goal $g$ and the goal-conditioned policy $\pi(.|g)$ that generated it can be written \cite{ho2019communication, baker2009action} as:

\begin{equation} \label{eq:traj_goal}
\begin{split}
\mathds{P}(d|g) & = \mathds{P}((s_1, a_1), .., (s_n, a_n)|g) \\
 & = \prod_{i=1}^n \pi(a_i|s_i,g) \cdot \mathds{P}(s_{i+1}|s_i, a_i) = \prod_{i=1}^n \pi(a_i|s_i,g),
\end{split}
\end{equation}

where $\mathds{P}(s_{i+1}|s_i, a_i) = 1$ as we consider a deterministic environment.

Now, to infer a goal given a demonstration, by using Bayes' rule we can derive $\mathds{P}(G|d)$, the probability distribution over the goal space $G$ given the demonstration:

\begin{equation} \label{eq:bayesian_tom}
\begin{split}
\mathds{P}(G|d) & \propto\mathds{P}(d|G) \cdot \mathds{P}(G)  = \prod_{i=1}^n \pi(a_i|s_i,G)  \cdot \mathds{P}(G).
\end{split}
\end{equation}

For each goal $g$, the prior $\mathds{P}(G)$ is uniform if not specified otherwise. Given $\mathds{P}(G|d)$, an agent can infer the goal of a demonstration by either taking the most probable goal, or sampling from the distribution. To perform this inference, the agent uses its own policy.

\paragraph{Learning Goal Inference from Own Trajectories}

When playing its policy, an agent produces trajectories which we call demonstrations when they are produced for other agents. A GCRL agent can leverage the {\sc bgi} mechanism on its own trajectories to improve its ability to infer goals. To do so, during training, the agent selects a goal, performs a trajectory, infers the goal from the trajectory, and rewards itself if the inference is correct. This reinforces the policy towards actions that lead to better goal inference. We use this to implement both pedagogy in the teacher and pragmatism in the learner.

\subsection{Naive/Pedagogical Teacher and Literal/Pragmatic Learner training}

We now present the two training phases in our teacher-learner setup. First, the teacher is pre-trained, and then it provides demonstrations for the learner to train with. The two-phases training process is presented in Algorithm~\ref{alg:teacher-learner}.

\begin{algorithm}[ht]
	\caption{Two-phases training of the teacher and the learner}
	\label{alg:teacher-learner}
	\begin{algorithmic}[1]
	    \State \textbf{PHASE 1: Teacher pre-training ($\pi_T(.|g)$)}
	    \State Initialize $\pi_T(.|g)$, goal set $G_T$ to $\{\text{first goal}\}$ and pedagogical Boolean variable
		\For {$epoch=1,2,\ldots$}
		    \State Randomly sample goal $g$ from goal set $G_T$
			\State Run policy $\pi_T(.|g)$, obtain trajectory $t=((s_1, a_1, r_1), .., (s_n, a_n, r_n))$ and achieved goal $g_a$
			\If{pedagogical Boolean $=1$ and $g=g_a$}
			    \State Infer own goal $\widehat{g}$ with {\sc bgi} on $t$: if $\widehat{g}=g$, add pedagogical reward to $t$ 
			\EndIf
			\State Add $t$ to replay buffer and if $g_a$ is new, add $g_a$ to goal set $G_T$
			\State Update policy $\pi_T$ with GCRL
		\EndFor

	\hrulefill
	\end{algorithmic} 
	
	\begin{algorithmic}[1]
	    \State \textbf{PHASE 2: Learner ($\pi_L(.|g)$) training with Teacher's demonstrations}
	    \State Initialize $\pi_L(.|g)$, goal set $G_L$ to $\{\text{first goal}\}$ and pragmatic Boolean variable
		\For {$epoch=1,2,\ldots$}
		    \State Teacher randomly samples goal $g_d$ from goal set $G_L$ and gives demo $d$ by running $\pi_T(.|g_d)$
		    \State Learner infers goal $\widehat{g_d}$ of $d$ with {\sc bgi}
		    \If{$\widehat{g_d} = g_d$ (feedback provided by the teacher)}
		        \State Add $d$ to learner's replay buffer
		    \EndIf
			\State Run policy $\pi_L(.|\widehat{g_d})$, obtain trajectory $t=((s_1, a_1, r_1), .., (s_n, a_n, r_n))$ and achieved goal $g_a$
			\If{pragmatic Boolean $=1$ and $\widehat{g_d}=g_a$}
			    \State Infer own goal $\widehat{g_o}$ with {\sc bgi} on $t$: if $\widehat{g_o}=g_a$, add pragmatic reward to $t$ 
			\EndIf
			\State Add $t$ to learner's replay buffer and if $g_a$ is new, add $g_a$ to goal set $G_L$
			\State Update policy $\pi_L$ with GCRL
		\EndFor
		
	\end{algorithmic} 
\end{algorithm}

\vspace{1cm}

\textbf{Phase 1: Teacher pre-training.} The teacher is pre-trained before providing demonstrations to the learner. The {\bf naive teacher}'s policy is trained to master all goals: it maintains a goal set $G_T$ to which it adds newly encountered goals, it samples goals from $G_T$ and pursues them to collect trajectories added to a replay buffer. Finally, it applies GCRL for policy updates.

The {\bf pedagogical teacher} is also trained to master all goals, but additionally trains itself to predict its own goals using {\sc bgi}. When training, each time it successfully reaches a goal, it takes its own trajectory and infers the pursued goal using {\sc bgi}, effectively asking itself: "would a learner be able to predict the goal from this demonstration?". If it correctly infers the goal from its own trajectory, it rewards itself to reinforce its probability to select this trajectory. This is done with GCRL by adding a "pedagogical reward" to the trajectory for which the teacher correctly infers its own goal. This biases policy learning towards finding demonstrations which avoid ambiguity in goal inference, while guaranteeing high performance on the actual task.

\textbf{Phase 2: Training the Literal/Pragmatic Learner with Teacher's demonstrations.} The {\bf literal learner}'s goal-conditioned policy $\pi_L(.|g)$ is trained to master all goals in the environment using teacher's demonstrations and GCRL.

At each iteration, the teacher samples a desired goal $g_d$ it wants the learner to achieve. It then presents a demonstration $d$ for the chosen goal to the learner. Using {\sc bgi} with its own policy, the learner infers the goal $\widehat{g_d}$ from the teacher's demonstration. The teacher provides feedback, telling the learner if $g_d = \widehat{g}$. If the learner correctly inferred the goal of the teacher's demonstration ($\widehat{g_d} = g_d$), the demonstration is added to the learner's replay buffer and used for further training. Then the learner plays its policy conditioned on the predicted goal $\pi_L (.|\widehat{g_d})$ and obtains a trajectory $t$ resulting in an achieved goal $g_a$, which is added to the replay buffer. If $g_a$ is achieved for the first time, then the teacher updates the set of goals from which it can sample. Finally, GCRL improves the policy from its own trajectories and demonstrations drawn from the replay buffer. Note that we assume that the learner does not have access to the goal of the demonstration and thus must infer it, mimicking real-life situations where humans regularly have to infer other people's goals \cite{gweon2021inferential}.

A {\bf pragmatic learner} improves over a literal learner by training its own policy to better infer the right goals from the teacher's demonstrations. If the learner is able to infer its own goals using {\sc bgi}, then it will better infer the teacher's goals. The pragmatic learner thus adds a "pragmatic reward" to its trajectories for which it can correctly infer its own goals.

\section{Experimental Setup}

We use two environments as test-beds for our experiments: a simple one inspired from research in Developmental Psychology \cite{gweon2021inferential} called "Draw two balls" (DTB) that we initially used for illustrating our approach, and a more complex one called "Fetch block-stacking" (FBS) to test our approach on a more challenging domain \cite{plappert2018multi}. Results on DTB are presented in Appendix~\ref{app:DTB}, and the main paper focuses on FBS. Both DTB and FBS are multi-goal environments and present goal ambiguity.

\subsection{Environment: Fetch Block Stacking}

\begin{wrapfigure}{r}{5cm}
\centering
\includegraphics[scale=0.15]{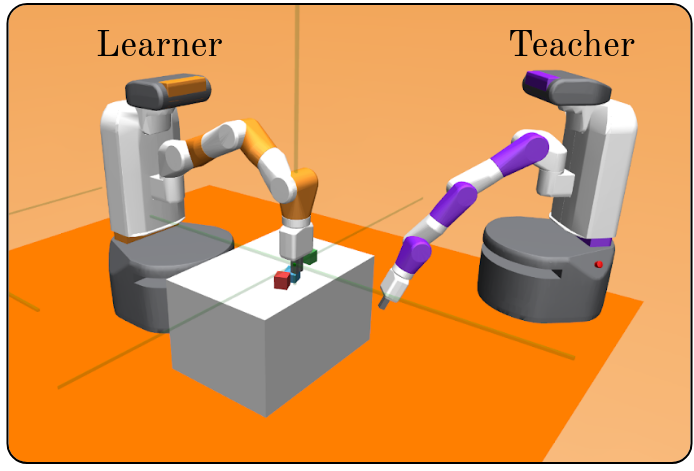}
\caption{Fetch Block Stacking.}
\label{fig:fetch}
\end{wrapfigure}

FBS is a block-stacking environment with two Fetch robots (teacher and learner) equipped with robotic arms, see \figurename~\ref{fig:fetch}. It is based on MuJoCo \cite{todorov2012mujoco} and derived from the Fetch tasks \cite{plappert2018multi}.

The teacher and learner share goal/state/action spaces. Actions are 4-dimensional: 3D gripper velocities and grasping velocity. Observations are the Cartesian and angular positions and velocities of the gripper and the three blocks. Following \cite{akakzia2021grounding}, we adopt a semantic predicates representation where we add to the observation a binary vector telling whether two blocks are close, and whether any block is on top of any other. The agent uses these binary vectors as goals (with 35 possible goals including stacks and pyramids). Given a particular goal vector configuration, the agent gets a $+1$ reward for each pair of blocks ($3$ in total) if true predicates about the pair of blocks are the same in the current vector and the goal vector. There are several possibilities to match the true predicates in the goal vector, which induces goal ambiguity. Details about the environment implementation are provided in Appendix~\ref{app:fbs}.

\textbf{Naive/Pedagogical teachers implementation.} The training procedure and policy architecture of the naive teacher are taken from GANGSTR \cite{akakzia2022help} which already implements Fetch Block Stacking. The GANGSTR agent perceives the low-level geometric states and the high-level semantic configurations. Following \cite{akakzia2022help}, we encode both object-centered and relational inductive biases in our architecture. We model both the agents policies and critics as Message Passing Graph Neural Networks \cite{gilmer2017neural}. We consider a graph of 3 nodes, each representing a single object. All the nodes are interconnected. We consider the agent’s body attributes as global features of the policy networks and both the agent’s body attributes and the actions as global features of the critic networks. 

The agent's goal-conditioned policy is trained using Soft Actor Critic (SAC) \cite{haarnoja2018soft}, a state-of-the-art RL algorithm, combined with Hindsight Experience Replay (HER) \cite{andrychowicz2017hindsight}. Additionally, the pedagogical teacher rewards itself with the "pedagogical reward" ($1$ here) when it correctly infers the goal of its successful trajectories. It does so for each of its collected trajectories during training. All architecture, training and hyperparameters details are provided in Appendix~\ref{app:details_fbs}.

\textbf{Literal/Pragmatic learners implementation.} To implement literal and pragmatic learner training, we use a combination of GANGSTR combined with a slightly modified version of the Soft Q-Imitation Learning (SQIL) algorithm \cite{reddy2019sqil}. Originally, SQIL rewards demonstrations to 1 and experience to 0. By contrast, we set the demonstration reward to $1$ + the maximum reward obtainable in the environment ($3$ here), while the reward of trajectories performed in the environment is unchanged. This allows the agent to learn from demonstrations and collected trajectories at the same time using SAC as the GCRL algorithm. As in the original paper, the percentage of demonstrations versus collected trajectories is set to $0.5$. On top of that, the pragmatic learner adds the "pragmatic reward" ($1$ here) to its trajectories from which it can infer the goal.

\subsection{Metrics}

We use four metrics for all of our experiments: 1) Goal Inference Accuracy ({\sc gia}): "is the learner able to correctly infer goals given demonstrations from a teacher?", 2) Own Goal Inference Accuracy ({\sc ogia}): "is the agent (teacher or learner) able to correctly infer goals from its own trajectories?", 3) Goal Reaching Accuracy ({\sc gra}): "is the learner able to reach all goals in the environment?", and 4) the product of {\sc gia} and {\sc gra} ({\sc gia}x{\sc gra}), interpreted as the ability of the learner to predict its goal given a demonstration from the teacher and then reach it. {\sc gia} and {\sc ogia} help us understand if the pedagogy and pragmatism mechanism actually work, and are computed using a test set of 50 demonstrations per goal generated by the teacher for {\sc gia} and by the same agent being tested for {\sc ogia} ($50*35=1750$ demonstrations in total). {\sc gra} allows us to evaluate the performance of agents and is computed by testing the agent on each goal 50 times. All demonstrations are long enough to reach the goal ($100$ timesteps in our case). We provide means $\mu$ and standard deviations over $5$ seeds and report statistical significance using a two-tail Welch’s t-test with null hypothesis $\mu_1 = \mu_2$, at level $\alpha= 0.05$ (noted by star markers in figures).

Finally, to compare pedagogical and naive demonstrations, we derive an Ambiguity Score: given two ambiguous goals that can be achieved simultaneously and a starting state, the ambiguous score is $1$ if their associated demonstrations achieve the same goals (the demonstrations are thus ambiguous), and $0$ otherwise. This score assumes that the demonstrator already knows how to achieve the goals. When it is computed from several situations, it evaluates the degree of ambiguity in the demonstrations. For more details on this score please refer to Appendix~\ref{app:ambiguity}.

\section{Experiments and Results}

We first study learning results from Phase 1 (naive/pedagogical teachers training) and Phase 2 (literal/pragmatic learners trained with teacher's demonstrations). We analyze the impact of demonstrations to accelerate learning, and then scrutinize two algorithmic choices in our approach: goal inference with {\sc bgi} and learning from demonstrations with SQIL. 

%We take the GANGSTR policy architecture and algorithm as they are and refer to the original paper \cite{akakzia2022help} for their justification.

\subsection{Experiments on Phase 1: Naive and Pedagogical Teachers}

\chugo{rajouter des références à la vidéo}

We verify that the pedagogical teacher can indeed better predict goals from its demonstrations compared to a naive teacher in the FBS environment. Quantitatively, the pedagogical teacher achieves an {\sc ogia} of $95.6\% \pm 0.1$ while the naive teacher achieves $83.4\% \pm 0.2$.
By training to infer its own goals from its demonstrations, it produces less ambiguous demonstrations. For a qualitative evaluation of this phenomenon, we implement the example in the introduction. We compare demonstrations from a common starting situation (red block close to green block, blue block away) and for two different goals: "Put blue block close to red" and "Put all three blocks close together". As a result, the naive teacher provides the same demonstrations (putting the blue block close to red), while the pedagogical teacher provides two different demonstrations to better disambiguate between the two goals (putting the red block close to blue and putting the blue block close to red and green). 

We compute the Ambiguity Score from 500 ambiguous situations sampled from a list of ambiguous situations detailed in Appendix~\ref{app:ambiguity} (two ambiguous goals and a starting state) and obtain an unequivocal result: $9\% \pm 1$ for the pedagogical teacher vs $64\% \pm 2$ for the naive teacher. Pedagogical demonstrations are thus 7 times less ambiguous than naive ones. 

%\chugo{les cas ou le pedagogique rate: simulation pas terrible ou alors rate la task. dire que le score d'ambiguité prend en comme hypothese que l'agent sait faire toutes les taches.}

\subsection{Experiments on Phase 2: Literal and Pragmatic Learners}
\label{sec:phase2}

\begin{figure}[ht]
    \captionsetup{justification=centering}
    \centering
    \includegraphics[scale=0.16]{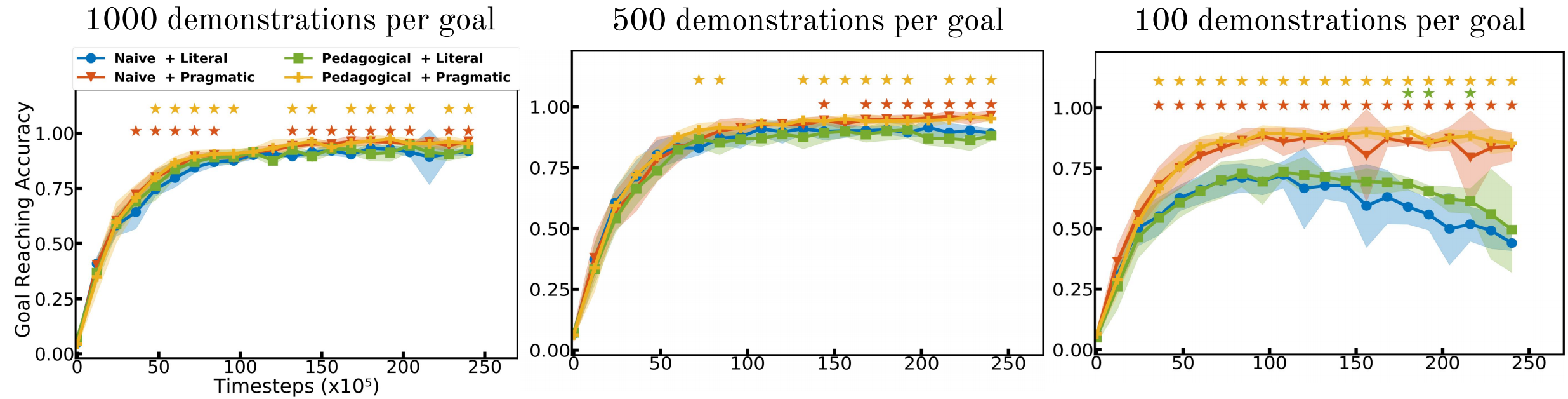}
    \caption{Global learner performance in the FBS environment (Goal Reaching Accuracy, GRA) with different numbers of demonstrations per goal. Table 1 shows that the drop in GRA under the few demonstrations regime (right) is mainly due to incorrect goal inference in the literal learner or with the naive teacher. Stars indicate significance (tested against naive+literal).}
    %\caption{Results for FBS environment (Goal Reaching Accuracy ({\sc gra}) with different numbers of demonstrations per goal). Stars indicate significance (tested against naive+literal).}
    \label{fig:results}
\end{figure}

We train pragmatic/literal learners with naive/pedagogical teachers and present our results in \figurename~\ref{fig:results} and Table~\ref{tab:results-fetch-all}. In order to evaluate how the mechanisms would respond to training regimes with few demonstrations, we performed the experiments in FBS with 1000, 500 and 100 demonstrations per goal, starting from random states. 

\pagebreak

As presented in \figurename~\ref{fig:results} and Table~\ref{tab:results-fetch-all}, it is highly preferable to opt for \bgi-agents (a pedagogical teacher with a pragmatic learner). The benefits of adopting pedagogy and pragmatism are particularly prominent in the few demonstrations regime (100 per goal) where pedagogical+pragmatic combination performs $4$ times better (going from $0.15$ to $0.62$) than the naive+literal one in terms of {\sc gia}x{\sc gra}. Note that with less than 100 demonstrations per goal, none of the approaches are able to master all goals, even if the hierarchy of methods remains unchanged (pedagogy and pragmatism being better than naive and literal, see \figurename~\ref{fig:10demos_fig} in Appendix~\ref{app:10demos}). This is likely due to SQIL forcing a 50/50 split of demonstrations and collected trajectories in the buffer.

An analysis of the pragmatic learner policies reveals that they are better at predicting their own goals, as intended, with an average of $12.4\%$ relative difference in {\sc ogia} between a literal and a pragmatic learner. This, coupled with the effectiveness of pedagogical demonstrations shown in the previous section, helps the teacher and learner efficiently communicate and improves learning speed. The benefit of using pragmatism is prominent in the few demonstrations regime: the addition of pragmatism alone (red curve in \figurename~\ref{fig:results}) helps mastering all goals while pedagogy on its own (green curve in \figurename~\ref{fig:results}) is not enough to do so. In additional experiments not reported here, we verified that the value of the pedagogical reward and pragmatic reward does not affect performance.

\begin{table}[ht]
  \captionsetup{justification=centering}
  \caption{Goal Reaching Accuracy ({\sc gra}) and Goal Inference Accuracy ({\sc gia}) for the FBS environment.}
  \label{tab:results-fetch-all}
  \centering
  \begin{tabular}{llll}
    \toprule
    Teacher + Learner    & {\sc gia}     & {\sc gra} & {\sc gia}x{\sc gra} \\
    \midrule
    \multicolumn{4}{c}{With 100 demonstrations per goal} \\
    \midrule
    Naive + Literal & $30.9 \pm 8.3\%$  & $50.2 \pm 2.7\%$ &  $0.15$   \\
    Naive + Pragmatic & $61.3 \pm 2.9\%$  & $84.8 \pm 4.1\%$ &  $0.52$   \\
    Pedagogical + Literal & $49.6 \pm 6.6\%$  & $62.7 \pm 5.4\%$ & $0.31$    \\
    Pedagogical + Pragmatic & $\mathbf{69.2 \pm 4.2\%}$  & $\mathbf{89.4 \pm 2.0\%}$ & $\mathbf{0.62}$     \\
    \midrule
    \multicolumn{4}{c}{With 500 demonstrations per goal} \\
    \midrule
    Naive + Literal & $68.9 \pm 1.1\%$  & $90.6 \pm 1.4\%$ &  $0.62$   \\
    Naive + Pragmatic & $76.3 \pm 0.4\%$  & $95.5 \pm 0.6\%$ &  $0.73$   \\
    Pedagogical + Literal & $78.8 \pm 0.7\%$  & $91.5 \pm 1.1\%$ & $0.72$    \\
    Pedagogical + Pragmatic & $\mathbf{82.2 \pm 1.2\%}$  & $\mathbf{96.8 \pm 0.6\%}$ & $\mathbf{0.80}$    \\
    \midrule
    \multicolumn{4}{c}{With 1000 demonstrations per goal} \\
    \midrule
    Naive + Literal & $75.3 \pm 0.9\%$  & $93.3 \pm 1.2\%$ & $0.70$    \\
    Naive + Pragmatic & $76.0 \pm 0.9\%$  & $96.6 \pm 0.4\%$ & $0.73$    \\
    Pedagogical + Literal & $80.0 \pm 1.6\%$  & $94.5 \pm 1.2\%$ & $0.76$    \\
    Pedagogical + Pragmatic & $\mathbf{84.7 \pm 1.8\%}$  & $\mathbf{97.1 \pm 1.2\%}$ & $\mathbf{0.82}$     \\
    \bottomrule
  \end{tabular}
\end{table}

\vspace{1cm}
\subsection{Does using the teacher accelerate training for the learner?}

\begin{wrapfigure}{r}{7.cm}
    \centering
    \includegraphics[scale=0.12]{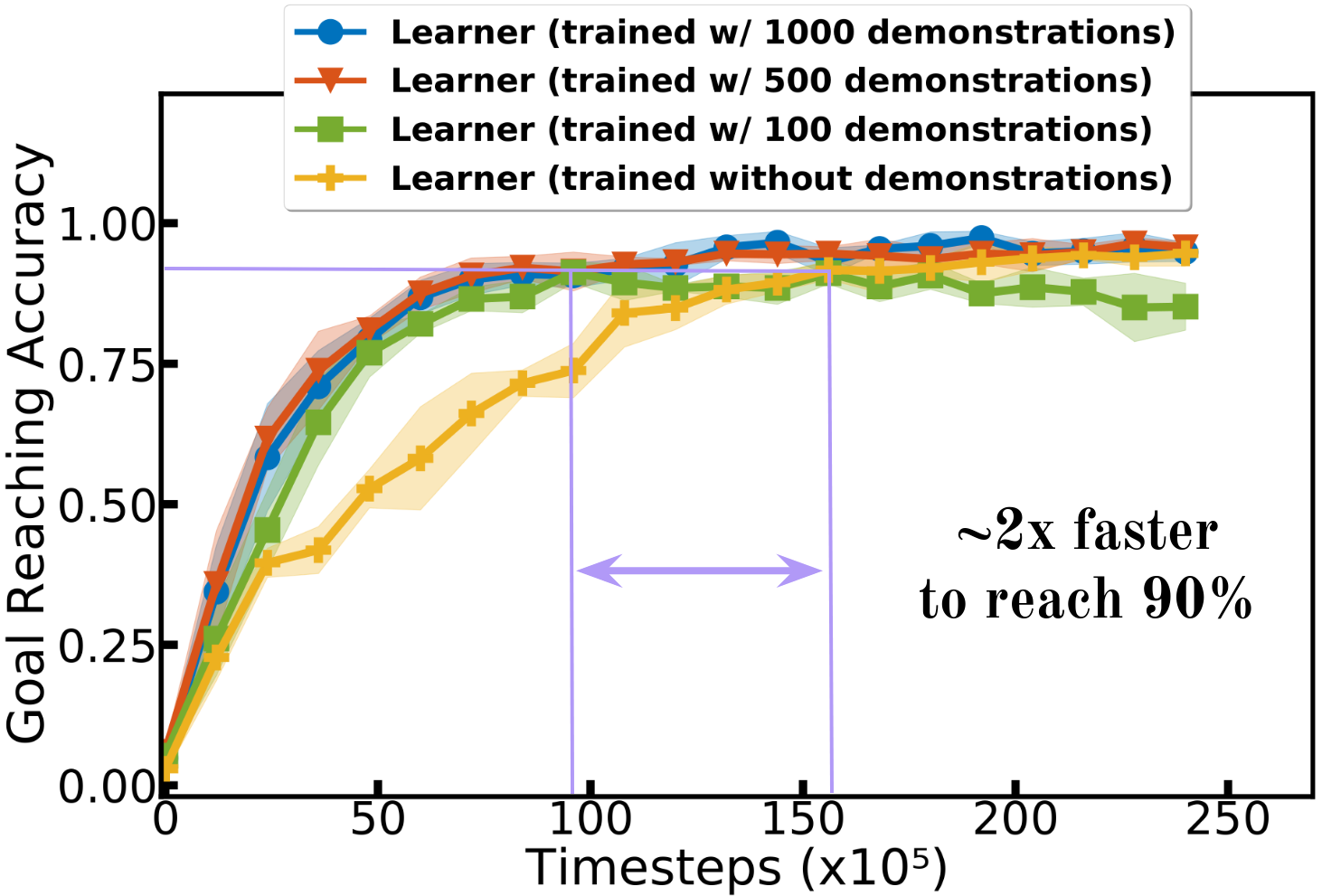}
    \caption{Demonstrations double the learner's training speed.}
    \label{fig:training_speed}
\end{wrapfigure}

In Phase 2 of our setup, the learner is trained using teacher's demonstrations and RL on its own experience. But is the teacher useful? The answer lies in the training speed. We thus compare the training speed of  pragmatic learners trained with $100$, $500$ and $1000$ pedagogical demonstrations and their own collected experience (using the same training procedure as before, with SQIL and GANGSTR), and a learner trained with the same architecture and training process, but no demonstration. We end up with a common result in the Learning from Demonstrations literature: the learner trains much faster (here roughly twice faster to obtain a {\sc gra} of $90\%$) when it has access to demonstrations, as illustrated in \figurename~\ref{fig:training_speed}. This sanity check justifies training learners from demonstrations.

\subsection{Could we use another goal inference method instead of Bayesian Goal Inference?}
\label{sec:gpnn}

\begin{table}[ht]
  \captionsetup{justification=centering}
  \caption{Goal Inference Accuracy ({\sc gia}) of {\sc bgi} and a GPNN on FBS.}
  \label{tab:bgi_ablation}
  \centering
  \begin{tabular}{llll}
    \toprule
    \multicolumn{4}{c}{{\sc gia} for Pedagogical demonstrations}\\
    \midrule
    Nb of demos per goal     & 100 demos     & 500 demos & 1000 demos \\
    \midrule
    GPNN & $46.2\%$  & $70.0\%$ & $82.0\%$    \\
    BGI & $\mathbf{69.2\%}$  & $\mathbf{82.2\%}$ &  $\mathbf{84.7\%}$   \\
    \midrule
   \multicolumn{4}{c}{{\sc gia} for Naive demonstrations}\\
    \midrule
    Nb of demos per goal     & 100 demos     & 500 demos & 1000 demos \\
    \midrule
    GPNN & $38.4\%$  & $67.1\%$ & $68.2\%$    \\
    BGI & $\mathbf{61.3\%}$  & $\mathbf{76.3\%}$ &  $\mathbf{76.0\%}$   \\
    \bottomrule
  \end{tabular}
\end{table}

Our learner relies on {\sc bgi} using its own policy to infer the goals associated with the demonstrations of the teacher. Could we instead train a separate goal prediction module to learn how to achieve goal inference? We tested this by creating a goal prediction neural network (GPNN) based on a LSTM architecture \cite{hochreiter1997long} that inputs a demonstration and outputs a goal, and we trained it using the same demonstrations and goals as for the {\sc bgi} case. In order to compare our {\sc bgi} approach to this alternative, we trained GPNN on a different number of demonstrations per goal and compared the results on a separate test set. We performed this experiment on FBS and provide details in Appendix~\ref{app:bgi}. Results from Table~\ref{tab:bgi_ablation} show that {\sc bgi} achieves higher {\sc gia} in all cases (naive or pedagogical demonstrations), and again performs especially well in the few demonstrations regime where the GPNN approach struggles because of too few training samples. However, note that in the case of an infinite goal space, a GPNN approach would be viable, while an approach based on {\sc bgi} would not be straightforward to apply.

%\pagebreak

\subsection{What if we used another method to learn from demonstrations?}

\begin{wraptable}{r}{7.5cm}
    \vspace{-0.3cm}
    \caption{Learning from demonstrations baselines.}
    \label{tab:baselines-demos}
    \centering
    \begin{tabular}{l|ccc}
    Method  &  {\sc gia} ($\%$)  & {\sc gra} ($\%$) & {\sc gia}x{\sc gra} \\
    \hline    
    B1 & $5.3 \pm 1.5$ & $6.2 \pm 0.7$ & $0.00$\\
    B2 & $2.3 \pm 1.8$ & $52.5 \pm 2.1$ & $0.00$\\
    B3 & $59.1 \pm 2.4$ & $73.8 \pm 1.7$ & $0.43$ \\
    Ours & $\mathbf{69.2\pm4.2}$  & $\mathbf{89.4\pm2.0}$ & $\mathbf{0.62}$  \\
    \end{tabular}
    \vspace{-0.3cm}
\end{wraptable}

Our modified version of SQIL helps the learner leverage both demonstrations and the experience it collects. We justify the use of this method by comparing it to baseline approaches: learning only from demonstrations (B1), adding behavioural cloning on the demonstrations (B2) and the original version of SQIL (B3). All implementation details are provided in Appendix~\ref{app:baselines}. We perform the comparison using teacher's pedagogical demonstrations and pragmatism in the learner for all methods. B1 is not able to learn, B2 is able to achieve a reasonable {\sc gra} score but struggles to predict goals from demonstrations because the behavioural cloning updates interfere with the {\sc bgi} mechanism by changing the action probabilities. B3 and our modified SQIL approach perform well, with a significant advantage for our approach. The original SQIL was designed for single goal environments and does not provide enough goal-reward associations, while our modification extends the approach to multi-goal environments by providing a reward signal for the collected experience.

\chugo{Generalization on different starting states.}

\section{Conclusion}
\label{sec:ccl}
In this paper, building on the pedagogy and pragmatism concepts from Developmental Psychology, we have shown how learning from demonstration can benefit from a Bayesian goal inference mechanism. Using \bgi-agents over regular agents improves learning speed, resolves ambiguity in communication and globally improves performance, especially in the few demonstrations regime.
In our work, the teacher's demonstration was insensitive to the learner's interpretation, as the teacher did not use a model of the learner's policy to provide a demonstration tailored to that specific learner. Conversely, the learner did not have a model of the teacher's policy which may help it interpret the goal the teacher wanted to convey. In the future, adding in each agent a model of the other agent's policy would result in closing the interaction loop and favoring the emergence of richer interaction mechanisms, such as both partners using signals to help the other update their model of each other during training.
Besides, our approach was focused on goal disambiguation for pedagogy and improving goal inference for pragmatism. Pedagogy could be further improved with a curriculum of goals that would improve exploration \cite{akakzia2022help}. Pragmatism could be improved with pedagogy detection \cite{gweon2021inferential} and inductive generalization \cite{gweon2010infants}. In the case of multiple teachers, the learner could also develop a mechanism to evaluate teachers and select the most suited one \cite{nguyen2012active}. We leave all this for future work.

Finally, a special trait of our learner is that it learns both from its own signals and from interaction with its teacher, in accordance with the guidelines of \cite{sigaud2021towards}. While pedagogy and pragmatism are mechanisms that improve how agents can be taught, they suggest a number of other mechanisms to do so, which in combination to our approach could be beneficial such as language guided learning, internalization, motivation regulation and observational learning.

%\section*{Acknowledgements} Anonymized for reviews.

%\section*{Acknowledgements} This work was performed using HPC resources from GENCI-IDRIS (Grant 2020-A0091011875).

%This  project  has  received  funding from European Union’s  Horizon 2020 ICT-48 research and innovation actions under grant agreement No 952026 (HumanE-AI-Net).

%\pagebreak

\section*{Acknowledgements}

This work has received funding from European Union’s Horizon 2020 ICT-48 research and innovation actions under grant agreements No 952026 (HumanE-AI-Net) and No 765955 (ANIMATAS). This work was performed using HPC resources from GENCI-IDRIS (Grant 2022-A0131013011).

\bibliography{bibli}
\bibliographystyle{plain}

%%%%%%%%%%%%%%%%%%%%%%%%%%%%%%%%%%%%%%%%%%%%%%%%%%%%%%%%%%%%
\section*{Checklist}

\begin{enumerate}

\item For all authors...
\begin{enumerate}
  \item Do the main claims made in the abstract and introduction accurately reflect the paper's contributions and scope?
    \answerYes{}
  \item Did you describe the limitations of your work?
    \answerYes{See Section~\ref{sec:ccl}.}
  \item Did you discuss any potential negative societal impacts of your work?
    \answerYes{This work proposes a general framework for implementing pedagogy and pragmatism in learning from demonstrations, which does not have extra negative societal impacts beyond learning from demonstrations.}
  \item Have you read the ethics review guidelines and ensured that your paper conforms to them?
    \answerYes{}
\end{enumerate}

\item If you are including theoretical results...
\begin{enumerate}
  \item Did you state the full set of assumptions of all theoretical results?
    \answerNA{}
	\item Did you include complete proofs of all theoretical results?
    \answerNA{}
\end{enumerate}

\item If you ran experiments...
\begin{enumerate}
  \item Did you include the code, data, and instructions needed to reproduce the main experimental results (either in the supplemental material or as a URL)?
    \answerYes{}
  \item Did you specify all the training details (e.g., data splits, hyperparameters, how they were chosen)?
    \answerYes{See Appendix~\ref{app:implementation_details}.}
	\item Did you report error bars (e.g., with respect to the random seed after running experiments multiple times)?
    \answerYes{See all figures and tables.}
	\item Did you include the total amount of compute and the type of resources used (e.g., type of GPUs, internal cluster, or cloud provider)?
    \answerYes{See Appendix~\ref{app:gcrl}.}
\end{enumerate}

\item If you are using existing assets (e.g., code, data, models) or curating/releasing new assets...
\begin{enumerate}
  \item If your work uses existing assets, did you cite the creators?
    \answerYes{}
  \item Did you mention the license of the assets?
    \answerYes{}
  \item Did you include any new assets either in the supplemental material or as a URL?
    \answerNo{}
  \item Did you discuss whether and how consent was obtained from people whose data you're using/curating?
    \answerNA{}
  \item Did you discuss whether the data you are using/curating contains personally identifiable information or offensive content?
    \answerNA{}
\end{enumerate}

\item If you used crowdsourcing or conducted research with human subjects...
\begin{enumerate}
  \item Did you include the full text of instructions given to participants and screenshots, if applicable?
    \answerNA{}
  \item Did you describe any potential participant risks, with links to Institutional Review Board (IRB) approvals, if applicable?
    \answerNA{}
  \item Did you include the estimated hourly wage paid to participants and the total amount spent on participant compensation?
    \answerNA{}
\end{enumerate}

\end{enumerate}

%%%%%%%%%%%%%%%%%%%%%%%%%%%%%%%%%%%%%%%%%%%%%%%%%%%%%%%%%%%%
\newpage
\appendix

%\section{TODO list pour l'appendix}

\chugo{donner le code}

\chugo{faire la vidéo et l'ajouter au papier}

\chugo{expés de rapidité d'inférence une fois que les learners sont entraînés}

\section{Experiments on toy environment}
\label{app:DTB}

\subsection{Draw Two Balls environment}

\begin{figure}[ht]
\centering
\includegraphics[scale=0.15]{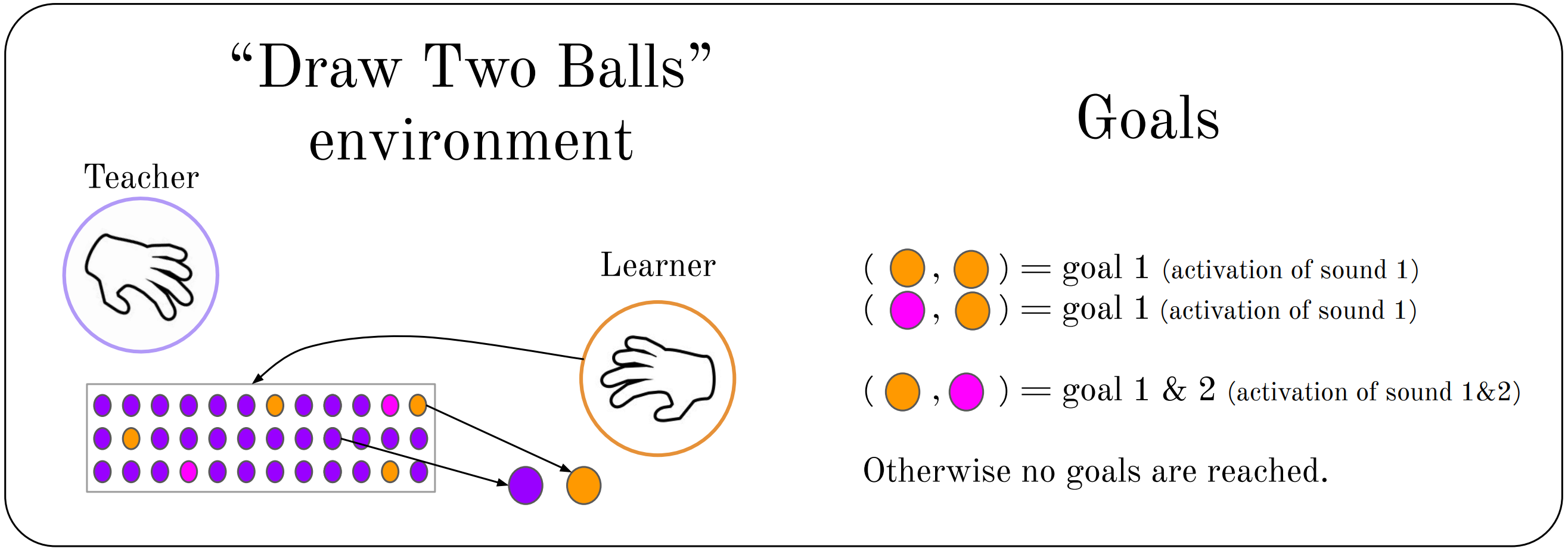}
\caption{The "Draw Two balls" (DTB) environment.}
\label{fig:dtb}
\end{figure}

The DTB environment is inspired from the squeaking balls environment presented in \cite{gweon2021inferential}, see \figurename~\ref{fig:dtb}. In our variant, there is a bucket of purple, orange and pink balls to choose from. Purple balls are more numerous than any other balls. The agent must pick two balls consecutively, and upon its choice, it can obtain three possible goals. If the picked balls are (orange, orange) or (pink, orange), goal 1 is reached. If the picked balls are (orange, pink), goal 1 and goal 2 are achieved. Otherwise, no goal is reached and nothing happens, which we will call reaching goal 0. The achievement of the goals are communicated to the agent via a characteristic sound is played upon the reaching of a goal as in \cite{gweon2021inferential}. Note that we purposely introduced goal ambiguity in the environment: if the agent selects the actions (orange, pink), one does not know which goal it was aiming for without an hypothesis on the agent. The DTB environment can then be adapted as a teacher-learner environment where the teacher can demonstrate the goals and the learner can infer the goals from the demonstrations.

We use the same two phases training process in DTB as in FBS, and describe thereafter the details of training on DTB.
\subsection{Phase 1: Teacher pre-training}

\textbf{Naive/pedagogical teacher implementation.} The naive teacher's policy $\pi_T$ is conditioned on the goal and parametrized by two probability distributions; the probability of selecting the first ball given the goal $g$: $\mathds{P}(\text{first ball} = x|g), x\in \{ orange, pink, purple\} $, and then the probability of selecting the second ball given the first ball and the goal $g$: $\mathds{P}(\text{second ball}=y | \text{first ball}=x,g)$. To get trained, the teacher randomly samples a goal and plays the corresponding policy using the conditional probabilities. The policy is then trained using a simple update rule: if the action sequence (pick the first and second ball) leads to the achievement of the pursued goal, we increase the probability of selecting this action sequence. Otherwise, we decrease this probability. For the pedagogical teacher, we increase even more the probability of an action sequence that reaches the goal and for which the teacher is able to infer the goal. 

\subsection{Phase 2: Training the Literal/Pragmatic Learner with Teacher’s demonstrations}

\textbf{Literal/pragmatic learner implementation.} The probability of selecting an action sequence is increased if the goal is correctly predicted and reached. As in FBS, pragmatism is implemented in DTB identically to the pedagogy mechanism of the teacher described above.

\subsection{Experiments and results}

\begin{figure}[ht]
    \centering
    \includegraphics[scale=0.2]{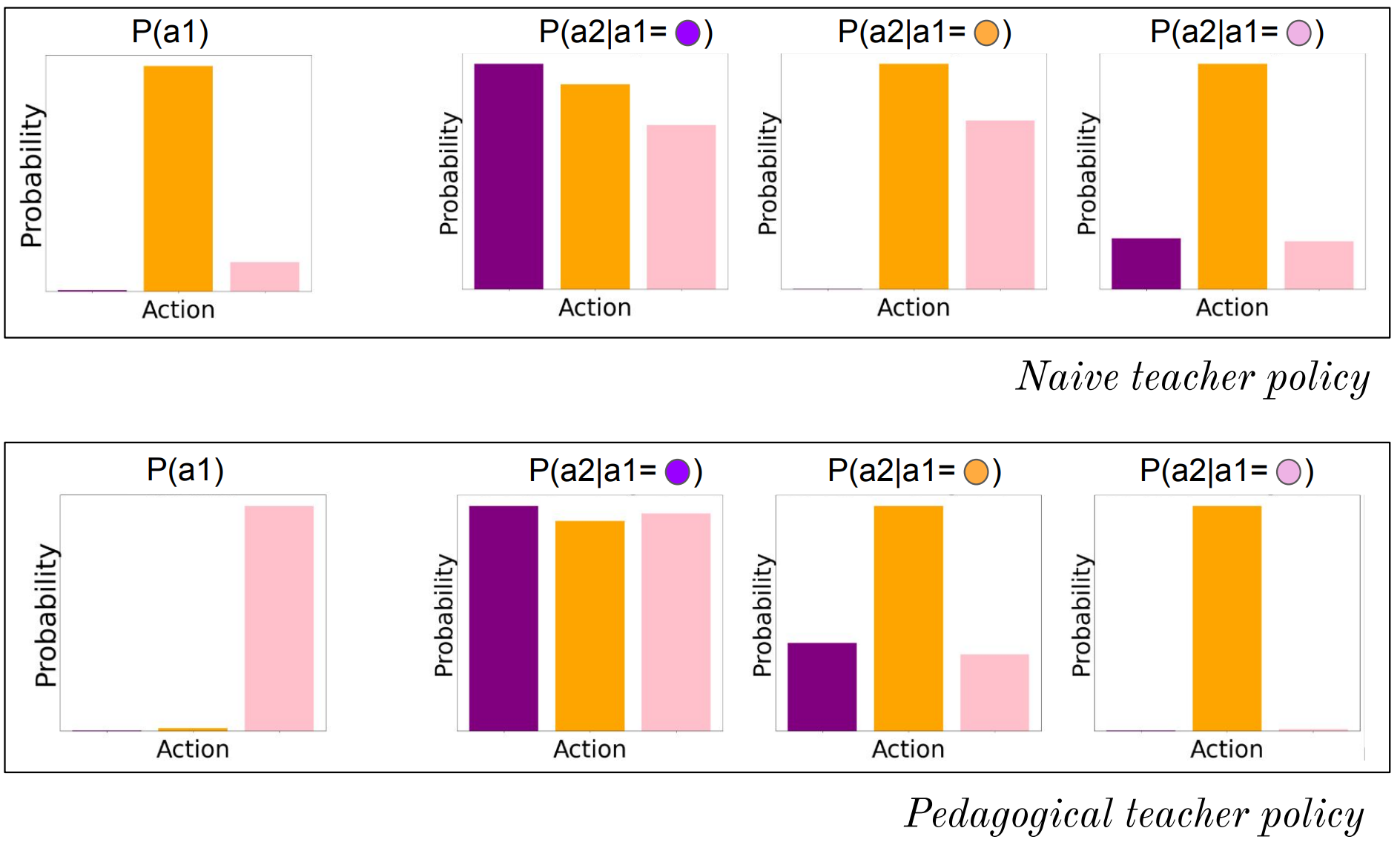}
    \caption{Analysis of naive vs pedagogical teachers on DTB for goal 1, which can be achieved by selecting either (pink, orange) or (orange, pink). Colored bars represent the probability of selecting the ball: on the left for the first action and on the right for the second action, given the first action. With a pedagogical demonstration, there is no ambiguity regarding the pursued goal: the pedagogical teacher always selects the unambiguous demonstration (pink, orange), while the naive teacher alternatively selects (pink, orange) or (orange, pink).}
    \label{fig:teachers_DTB}
\end{figure}

\paragraph{Phase 1: Qualitatively, what is the difference between a pedagogical and a naive demonstration?}

In order to answer this question, we analyze the teachers' policies. \figurename~\ref{fig:teachers_DTB}, presents the teacher policies for goal 1, illustrating the difference between a naive and a pedagogical teacher on DTB environment. Contrary to the naive teacher, the pedagogical teacher specifically avoids the demonstrations (orange, orange) for goal 1 because it does not allow to directly detect goal 1 after the first ball is picked. Moreover, it avoids the ambiguous demonstration (orange, pink), which reaches both goal 1 and goal 2. By carefully selecting among all possible demonstrations for a goal, this pedagogical teacher policy maximally avoids ambiguity in demonstrations. Quantitatively, this results in a Own Goal Inference Accuracy ({\sc ogia}) of $82\%$ for the naive teacher, and $100\%$ for the pedagogical teacher.

\begin{figure}[ht]
    \centering
    \includegraphics[scale=0.23]{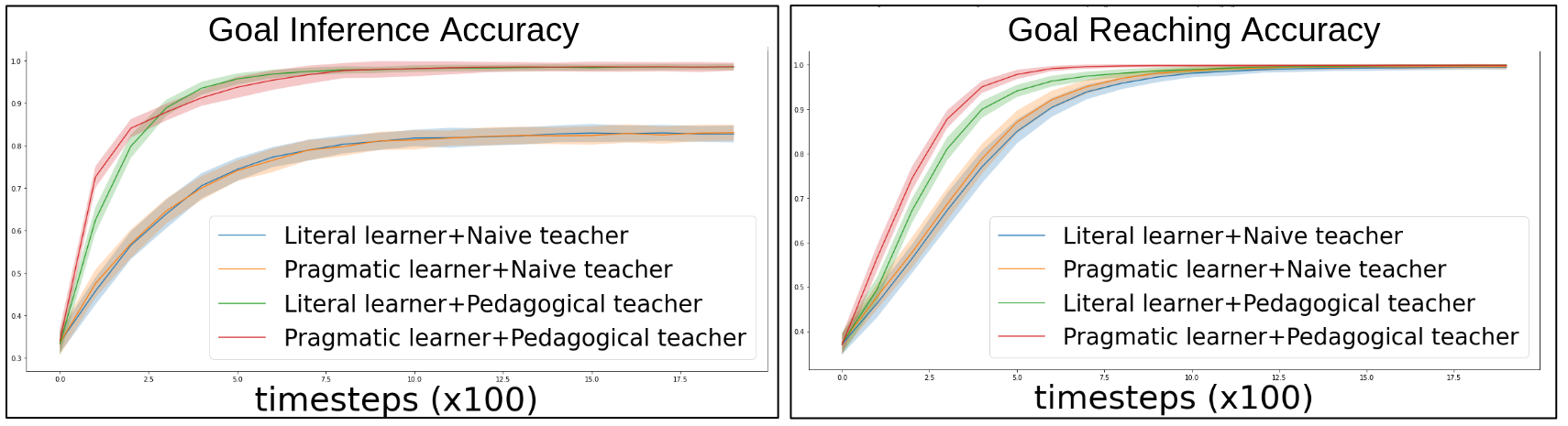}
    \caption{Left: Goal Inference Accuracy evaluated during training. Right: Goal Reaching accuracy evaluated during training.}
    \label{fig:learners_DTB}
\end{figure}

\paragraph{Phase 2: What are the benefits and drawbacks of using a pedagogical teacher over a naive teacher? A pragmatic learner over a literal learner?}

We experiment with pedagogical/naive tutors and pragmatic/literal learners just as in FBS, and present our results on \figurename~\ref{fig:learners_DTB}. When a learner is trained with a pedagogical tutor, it consistently learns faster, and can predict the goals from all demonstrations, whereas it is not capable of disambiguating goal 1 from goal 2 with the (orange, pink) demonstration from a naive tutor. Moreover, a learner benefits from pragmatism if the tutor is pedagogical, resulting in the best tutor-learner combination. The pedagogical teacher + pragmatic learner combination reaches a {\sc gra} of $100\%$ twice faster than the naive teacher + literal learner combination. Furthermore, the pedagogical+pragmatic combination reaches a {\sc gra x gia} of $1$ while the naive+literal combination only reaches $0.82$. The conclusions from these results are thus the same as with FBS. Considering the difference of environments, policy architectures and training processes between the experiments on DTB and FBS, it confirms that pedagogy and pragmatism do generally improve learning from demonstrations.

\section{Implementation details}
\label{app:implementation_details}

\subsection{Fetch Block Stacking environment}
\label{app:fbs}

As in \cite{akakzia2021grounding}, the environment goals are based on two predicates: the close and the above binary predicates. For the 3 objects we consider, these predicates are applied to all permutations of object pairs: 6 permutations for the above predicate and 3 combinations for the close predicate due to its order-invariance. A semantic configuration is the concatenation of these 9 predicates and represents spatial relations between objects in the scene. In the resulting semantic configuration space $\{0, 1\}^9$, the agent can reach 35 physically valid configurations, including stacks of 2 or 3 blocks and pyramids.

To compute the reward, the agent in FBS compares its goal configuration to the current configuration and derives a reward with the following procedure: it adds $1$ to the reward for each pair of blocks if the true predicates in the goal configuration match the current configuration. This means that the reward can either be $0$, $1$ or $3$ ($2$ is not achievable by associativity).

FBS is based on Mujoco which is licensed under the Apache License 2.0.

\subsection{Architecture, training, and hyperparameters for the teacher and learner}
\label{app:details_fbs}

\subsubsection{Object-centered architecture}

Both the teacher and learner share the same architectures, goal-conditioned RL training and hyperparameters. They are all based on the GANGSTR agent \cite{akakzia2022help} which is a graph-based goal-conditioned RL agent capable of learning to master all goals on the Fetch Block Stacking environment. A single forward pass through this graph consists in three steps:

\begin{itemize}
    \item Message computation is performed for each edge.
    \item Node-wise aggregation is performed for each node.
    \item Graph-wise aggregation is performed once for all the graph.
\end{itemize}

We use max pooling for the node-wise aggregation and summation for the graph-wise aggregation.

The object-centered architecture uses two shared networks, $NN_{edge}$ and $NN_{node}$, respectively for the message computation and node-wise aggregation. Both are 1-hidden-layer networks of hidden size 256. Taking the output dimension to be equal to 3x the input dimension for the shared networks showed the best results. All networks use ReLU activations and the Xavier initialization. We use the Adam optimizer \cite{kingma2014adam}, with a learning rate $10^{-3}$. The list of hyperparameters is provided in Table~\ref{tab:hyperparm}.

\subsubsection{Hyperparameters}

\begin{table}[htbp!]

%\vspace{-5cm}
    \centering
    \caption{Hyperparameters.}
    \label{tab:hyperparm}
    \vspace{0.2cm}
    \begin{tabular}{l|c|c}
        Hyperparam. &  Description & Values. \\
        \hline
        $nb\_mpis$ & Number of workers & $24$ \\
        $nb\_cycles$ & Number of repeated cycles per epoch & $50$ \\
        $nb\_rollouts\_per\_mpi$ & Number of rollouts per worker & $10$ \\
        $rollouts\_length$ & Number of episode steps per rollout & $40$ \\
        $nb\_updates$ & Number of updates per cycle & $30$ \\
        $replay\_strategy$ & HER replay strategy & $final$ \\
        $k\_replay$ & Ratio of HER data to data from normal experience & $4$ \\
        $batch\_size$ & Size of the batch during updates & $256$ \\
        $\gamma$ & Discount factor to model uncertainty about future decisions & $0.99$ \\
        $\tau$ & Polyak coefficient for target critics smoothing & $0.95$ \\
        $lr\_actor$ & Actor learning rate & $10^{-3}$ \\
        $lr\_critic$ & Critic learning rate & $10^{-3}$ \\
        $\alpha$ & Entropy coefficient used in SAC  & $0.2$ 
    \end{tabular}
\end{table}

\subsubsection{Goal-conditioned RL training}
\label{app:gcrl}

The RL training procedure relies on SAC \cite{haarnoja2018soft} for the RL and HER \cite{andrychowicz2017hindsight} for goal relabelling. It uses the Message Passing Interface \cite{dalcin2011parallel} to exploit multiple processors. Each of the 24 parallel workers maintains its own replay buffer of size $10^6$ and performs its own updates. Updates are summed over the 24 actors and the updated actor and critic networks are broadcast to all workers. Each worker alternates between 10 episodes of data collection and 30 updates with batch size 256. To form an epoch, this cycle is repeated 50 times and followed by the offline evaluation of the agent. Each agent is trained for 100 epochs, totalling $24*10^6$ timesteps. Each training is performed $10$ times with 10 random seeds and results report error bars (standard deviation).

As for the particular case of the learner, we provide additional details: 

\begin{itemize}
    \item Our implementation of SQIL \cite{reddy2019sqil} uses 50\% experience and 50\% demonstrations in the replay buffer. 
    \item To generate demonstrations, the teacher randomly selects a starting state and makes sure that the goal is achieved.
    \item When training the learner, the goal demonstrated by the teacher is selected randomly among the goals discovered by the learner.
\end{itemize}

Regarding pedagogy and pragmatism, the extra reward ("pedagogical reward" and "pragmatic reward") is set to $1$.

\subsubsection{BGI implementation details}
\label{app:bgi}

\textbf{Computation with continuous action policy.} The {\sc bgi} computation is performed using the policy of the agent. In the case of Fetch Block Stacking and SAC, the actions are continuous. The policy outputs a goal-conditioned normal distribution with as many dimensions as the dimension of the action space. This distribution is sampled for goal-conditioned action selection. Given a vector of actions and the goal-conditioned normal distribution of actions of a policy, one can compute the probability of observing such action given each goal, and construct a probability distribution over the goal space of observing such action. This is done using the cumulative distribution function of the Normal distribution (probability of observing a value given the parameters of the distribution). 

We generalize this computation to a trajectory rather than a single action by taking the product of probabilities over the goal space and normalizing to a probability distribution.

\textbf{Goal Prediction Neural Network details.} In Sec.\ref{sec:gpnn}, we experiment with a Goal Prediction Neural Network that is trained offline to predict goals from demonstrations using a training dataset of demonstration provided by the teacher, in order to provide a comparison to {\sc bgi} inference. This neural network takes a demonstration as input to a LSTM \cite{hochreiter1997long} layer with $512$ hidden units, from which the output is fed to a fully connected layer. The final outputs are goal probabilities. The predicted goal of the demonstration is the one with the higher probability. The activation functions are ReLU. The GPNN is trained until convergence ($300$ epochs) and then tested, just like BGI, on a separate test set of $500$ demonstrations, from which we report the results in Tab.\ref{tab:bgi_ablation}. We use a batch size of $256$ and Adam \cite{kingma2014adam} as the optimizer, with a learning rate $10^{-3}$.

\subsubsection{Baselines of learning from demonstrations}
\label{app:baselines}

\textbf{Baseline 1 (B1).} For this baseline, we discard the experience collected by the learner in the replay buffer and only use the demonstrations provided by the teacher to perform the same learning procedure as the main experiment in Sec.\ref{sec:phase2}.

\textbf{Baseline 2 (B2).} In this experiment, we use the same learning procedure as in the main experiment in Sec.\ref{sec:phase2}, but additionally perform a L2 regularization on the output action probabilities by using the demonstrations as a target. This is done using a Mean Squared Error loss and Adam optimizer with a learning rate of $10^{-3}$ and a batch size of $256$.

\textbf{Baseline 3 (B3).} This baseline is the original version of SQIL \cite{reddy2019sqil}, please refer to their paper for implementation details.

\subsubsection{Ambiguity Score details}
\label{app:ambiguity}

We provide additional details about the Ambiguity Score and its computation. We manually created a list of situations from which ambiguity in demonstrations exists. These situations are composed of a starting state (the initial state of relations between blocks before the demonstration begins), and two potentially ambiguous goals. The teacher then performs a demonstration for each of the two goals, with the same initial state. If both demonstrations achieve the same goal (even though they were aiming for different goals), then they are considered ambiguous.

The list of ambiguous situations we use to compute the results in the paper is the following:

\begin{itemize}
    \item Initial state: green is close to red, blue further apart. Ambiguous goals: green is close to red + blue is close to green and green is close to blue + red further apart.
    \item Initial state: green is close to red, blue further apart. Ambiguous goals: green is close to red + blue is above green and blue is above green + red further apart.
    \item Initial state: green is above red, blue further apart. Ambiguous goals: green is above red + blue is close to red and blue is close to green + green further apart.
    \item Initial state: green is above red, blue further apart. Ambiguous goals: green is above red + blue is close to red and blue is close to red + green further apart.
    \item Initial state: green is above red, blue further apart. Ambiguous goals: green is above red + blue is close to green and blue is close to green + red further apart.
    \item Initial state: green is close to red, blue further apart. Ambiguous goals: green is close to red + blue is above green and red in a pyramid and blue is close to red and green + green is close to red.
\end{itemize}

Note that these ambiguous situations are augmented with all possible permutations of block colors.

\newpage

\subsubsection{Additional results with less than 100 demonstrations per goal}
\label{app:10demos}

\begin{figure}[ht]
    \centering
    \includegraphics[scale=0.28]{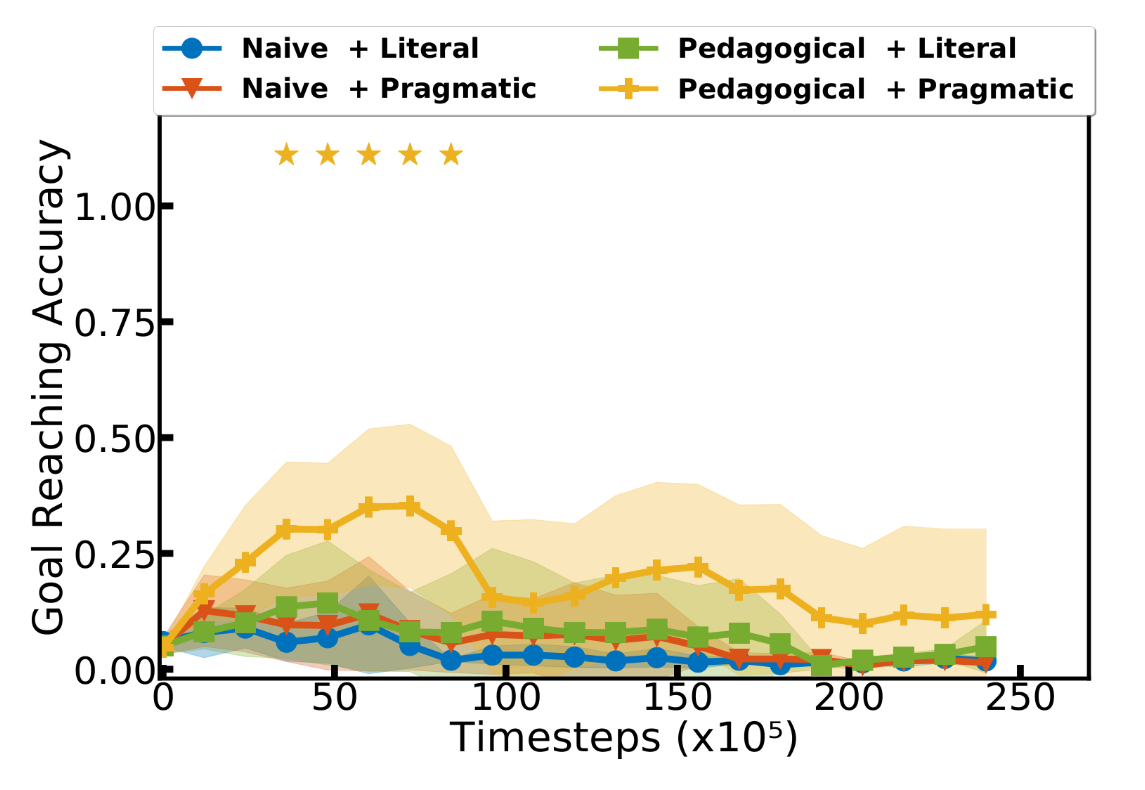}
    \caption{Results for FBS environment (Goal Reaching Accuracy ({\sc gra}) with 10 demonstrations per goal). Stars indicate significance (tested against naive+literal).}
    \label{fig:10demos_fig}
\end{figure}

The results with 10 demonstrations per goal show that none of the approaches are able to master all goals. However, the best combination is still a pedagogical teacher with a pragmatic learner, as \figurename~\ref{fig:10demos_fig} shows.

\end{document}